\pdfoutput=1

\documentclass[11pt]{article}

\usepackage{acl}

\usepackage{times}
\usepackage{latexsym}

\usepackage[T1]{fontenc}

\usepackage[utf8]{inputenc}

\usepackage{microtype}

\usepackage{graphicx}
\usepackage{geometry}
\usepackage{listings} 
\usepackage{url}
\usepackage{microtype}
\usepackage{amsmath,amssymb}
\usepackage{times}
\usepackage{latexsym}
\usepackage{graphicx}
\usepackage{multirow}
\usepackage{booktabs} 
\usepackage{tablefootnote}
\usepackage{colortbl} 
\usepackage{paralist} 
\usepackage{soul} 
\usepackage{subfigure}
\usepackage{xspace}
\usepackage{stfloats}

\definecolor{codegreen}{rgb}{0,0.6,0}
\definecolor{codegray}{rgb}{0.5,0.5,0.5}
\definecolor{codepurple}{rgb}{0.58,0,0.82}
\definecolor{backcolour}{RGB}{251, 253, 253}
\definecolor{codeorange}{RGB}{206, 127, 71}

\lstdefinestyle{mystyle}{
    backgroundcolor=\color{backcolour},   
    commentstyle=\color{codegreen},
    keywordstyle=\color{codepurple},
    numberstyle=\tiny\color{codegray},
    stringstyle=\color{codeorange},
    basicstyle=\ttfamily\footnotesize,
    breakatwhitespace=false,         
    breaklines=true,                 
    captionpos=b,                    
    keepspaces=true,                 
    numbers=left,                    
    numbersep=5pt,                  
    showspaces=false,                
    showstringspaces=false,
    showtabs=false,                  
    tabsize=2
}

\title{OpenICL: An Open-Source Framework for In-context Learning}

\author{
Zhenyu Wu$^{\blacklozenge \dagger *}$, 
Yaoxiang Wang$^{\clubsuit \dagger}$\thanks{\, Work done while interning at Shanghai AI Lab.}\;, 
Jiacheng Ye$^{\spadesuit}$\thanks{\, Equal Contribution.} \\
\textbf{Jiangtao Feng}$^{\diamondsuit}$,
\textbf{Jingjing Xu}$^{\diamondsuit}$,
\textbf{Yu Qiao}$^{\diamondsuit}$,
\textbf{Zhiyong Wu}$^{\diamondsuit}$\thanks{\, Corresponding Author.}
\\
$^\diamondsuit$Shanghai AI Laboratory \quad
$^\blacklozenge$ East China Normal University \\
$^\clubsuit$Xiamen University \quad
$^{\spadesuit}$The University of Hong Kong \\
\texttt{carsonye@cs.hku.hk, \{wuzhenyu,wangyaoxiang\}@pjlab.org.cn} \\
\texttt{\{fengjiangtao,xujingjing,qiaoyu,wuzhiyong\}@pjlab.org.cn}
}

\begin{document}
\maketitle
\begin{abstract}
In recent years, In-context Learning (ICL) has gained increasing attention
and emerged as the new paradigm for large language model (LLM) evaluation. Unlike traditional fine-tuning methods, ICL instead adapts the pre-trained models to unseen tasks \textit{without} any parameter updates.
However, the implementation of ICL is sophisticated due to the diverse retrieval and inference methods involved, as well as the varying pre-processing requirements for different models, datasets, and tasks. A unified and flexible framework for ICL is urgently needed to ease the implementation of the aforementioned components.
To facilitate ICL research, we introduce OpenICL, an open-source toolkit for ICL and LLM evaluation. OpenICL 
is research-friendly with a highly flexible architecture that users can easily combine different components to suit their needs. 
It also provides various state-of-the-art retrieval and inference methods to streamline the process of adapting ICL to cutting-edge research.
The effectiveness of OpenICL has been validated on a wide range of NLP tasks, including classification, QA, machine translation, and semantic parsing. As a side-product, we found OpenICL to be an efficient yet robust tool for LLMs evaluation. OpenICL is released at \url{https://github.com/Shark-NLP/OpenICL}.
\end{abstract}

\section{Introduction}
The rise of large language models~(LLMs)~\cite{tom2020gpt3, zhang2022opt, 	scao-2022-bloom} has shown impressive emergent In-Context Learning~(ICL) ability~\cite{wei2022emergent}.
Different from finetuning which requires parameter updates, ICL can perform inference with model parameters frozen.
ICL sidesteps the resource-intensive nature of fine-tuning, yet still yields comparable results to fine-tuned models in specific tasks~\cite{zhao2021calibrate,lu2022fantastically,gao2021making}. 
However, we observed a lack of a unified framework for ICL. Implementations from existing projects are often high-customized to their own needs, thus making further development and comparisons with previous approaches a challenge. 

The basic ICL pipeline contains two steps: retrieval and inference. Given a testing input $X'$, in the retrieval stage, several examples from the training set are retrieved as in-context demonstrations. In the inference stage, these demonstrations are prepended before $X'$ and fed into the LLM to generate the prediction. 
Researchers have explored various methods for both retrieval(e.g., BM25~\cite{PRF2009}, TopK~\cite{liu2022makes, gao2021making} and VoteK~\cite{su2022selective}) and inference(e.g., perplexity-based~\cite{tom2020gpt3}, channel-based~\cite{min2022channel}, and Chain-of-thoughts~\cite{wei2022cot}). However, these methods are often implemented under different frameworks, and/or evaluated using different LLMs and tasks. These inconsistencies make systematic evaluations and comparisons of various methods challenging, thus hindering the development of better ICL methods. 

To address this issue, we present OpenICL, an open-source and easy-to-use toolkit for ICL. OpenICL has many state-of-the-art retrieval and inference methods built in to facilitate systematic comparison and fast research prototyping.  OpenICL also provides a unified and flexible interface for the development and evaluation of new ICL methods. 
Users can easily incorporate different retrieval and inference methods, as well as different prompt instructions, into their pipelines. 
To validate OpenICL's implementation and design, we use OpenICL to evaluate LLMs on several NLP tasks, including classification, question answering, translation, and semantic parsing.  Our contributions are summarized as follows:

\begin{itemize}
    \item We propose OpenICL, an easy-to-use and extensible ICL framework for zero-/few-shot evaluation of language models
    \item OpenICL provides a wide range of ICL methods, LLMs, and tasks, requiring as little as a few lines of code to use and paving the way for more extensions in the future.
    \item We provide complete tutorials to walk users through the framework, thus facilitating research and development of ICL.
\end{itemize}

\section{Related Work}

\paragraph{In-context Learning} Besides the classic ``pre-train and fine-tune'' paradigm, \citet{tom2020gpt3} proposed In-context learning (ICL), a new paradigm that leverages pre-trained language models to perform new tasks without any gradient-based training. It appends a small number of training examples as prompts before the test input, and have shown to be able to improve LLMs’ performance in few-shot scenarios and generalize to a wide range of downstream tasks, such as information retrieval~\cite{informationretrieval}, fact checking~\cite{factchecking}, commonsense reasoning~\cite{commonsensereasoning}, arithmetic reasoning~\cite{cobbe2021gsm8k}, machine trainslation~\cite{agrawal2022selection, xi2021few-shot}, and data generation~\citep{ye-etal-2022-progen}, etc. 

Aside from those early successes, researchers have developed more sophisticated ICL methods that require some intermediate reasoning steps. Among them, chain-of-thoughts~(CoT) is the first attempt that significantly surpasses the previous state-of-the-art methods on many reasoning tasks~\cite{wei2022cot}. After that, different variants of CoT have been proposed to strengthen its performance, such as Self-Ask~\cite{selfask}, iCAP~\cite{wang2022iteratively}, Least-to-Most prompting~\cite{least}, and Selection-Inference~\cite{autocot, fu2022complexitycot}.

Despite the surprising performance, ICL has been criticized for being very sensitive to the choice and ordering of in-context examples~\cite{zhao2021calibrate,lu2022fantastically}. To address this problem, different criterion and context construction methods have been proposed. \citet{gao2021making} and \citet{liu2022makes} select examples that are closer to the test input in the embedding space; a line of work~\citep{su2022selective,levy2022diverse,ye2023compositional} select the most representative examples in the training set to encourage diversity of in-context examples; \citet{wu2022adpative} observe that Minimum Description Length~(MDL) principle can be an effective criterion for in-context example selection. 

\paragraph{Prompt Learning} Prompt learning~\cite{liu2021prompt} is a special case of ICL without any in-context examples. 
Prompt learning comprises various topics including manual template engineering~\citep{petroni-etal-2019-language,tom2020gpt3}, automated template learning~\citep{wallace-etal-2019-universal,shin-etal-2020-autoprompt,li-liang-2021-prefix}, and answer engineering~\citep{gao-etal-2021-making,schick-schutze-2021-just}. 
We refer the readers to the usage of OpenPrompt~\cite{openprompt} which is a toolkit specially designed for prompt learning. 
In comparison, OpenICL focuses more on integrating various exemplar retrieving approaches and inference strategies for in-context learning. Note that OpenICL can also seamlessly support prompt learning by setting the number of in-context examples to zero and specifying the manual or pre-searched prompt templates by OpenPrompt for different tasks.

\section{OpenICL}
In this section, we first explain OpenICL’s design principles. Then, we will briefly describe OpenICL’s two major components, namely, the \texttt{Retriever} and \texttt{Inferencer}.

\subsection{Design Principles} The design principle of OpenICL is to facilitate in-context learning research and enable efficient and robust large language model evaluation. In detail, we consider the following principles:
\paragraph{[P1: Modularity]} Since ICL is a fast-evolving research field, the design of OpenICL should be decoupled such that different components can be easily modified to support latest methods and/or combined to suit various tasks and application needs. 
\paragraph{[P2: Efficiency]} Nowadays, large language models can have hundreds of billions of parameters. To support inference at such a massive scale, OpenICL should be optimized to enable efficient parallel inference. 
\paragraph{[P3: Generality]} ICL has been widely used in all fields in NLP, so OpenICL needs a flexible interface that enables it to work with various LLMs, tasks, retrieval methods, and inference approaches.

\subsection{Architecture Overview}
\begin{figure*}[htbp]
    \centering
    \includegraphics[width=\textwidth]{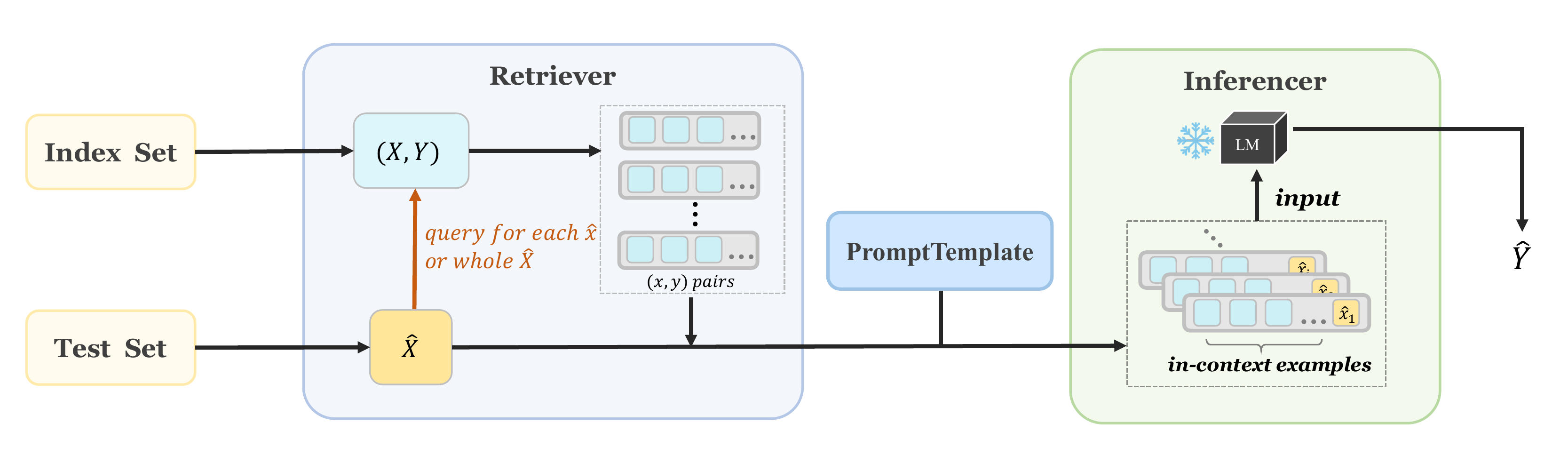}
    \caption{Overview of the architecture in OpenICL. OpenICL first obtains proper in-context examples from an index set for each test input or for the whole test set via retrieval methods (e.g., TopK or VoteK) specified by the users. Then the in-context examples and test input are concatenated into a single sequence based on the provided prompt template. Finally, all the prompts are fed into the language model to infer the output through defined inference strategies (e.g., Chain-of-thought).}
    \label{fig:archi}
\end{figure*}

Figure~\ref{fig:archi} overviews OpenICL's architecture. For each input $\hat{x}$ from the test set $\hat{X}$, the \texttt{Retriever} retrieves several $(x, y)$ pairs (represented as one row in the dashed box) from an index set $(X, Y)$ as $\hat{x}$'s in-context examples. These examples, as well as $\hat{x}$, are then formatted according to the user-defined prompt template and concatenated to form a text sequence. After that, the \texttt{Inferencer} digests these sequences and fed them into the LLMs to obtain the model prediction $\hat{Y}$. 

\subsection{Modularity}
To satisfy Principle P1, OpenICL adopts a loosely-coupled design between components. These components separate the data pre-processing, retrieval, and inference processes with very flexible interfaces that allow easy customization to fit specific needs. Two major components are detailed below: 

\paragraph{Retriever}
\verb|Retriever| is responsible for retrieving in-context examples from the pre-existing training data. This module supports both corpus-level (i.e., only retrieving one group of examples for the whole test set) and instance-level (i.e., retrieving examples for each testing input individually) retrieval methods. OpenICL primarily supports learning-free retrieval methods as follows:
\begin{itemize}
    \item Random: Early practice~\cite{tom2020gpt3} of ICL often randomly select examples to construct the context. Although Random brings high variance for ICL performance, it is still the popular choice when there are only a few demonstrations available~\cite{wei2022cot,calibrate}.
    \item Heuristic method: To overcome the disadvantage of Random, various semantic similarity based retrieval methods have been proposed and shown great promise, such as  BM25~\cite{PRF2009}, TopK~\cite{liu2022makes, gao2021making}, and VoteK~\cite{su2022selective}. 
    \item Model-based method: More recently, researchers have explored using models' confidence in the output to select and order examples, such as entropy~\cite{lu2022fantastically} and MDL~\cite{wu2022adpative}. 
\end{itemize}
OpenICL has implemented the existing methods above to facilitate future research and systematic comparison. Furthermore, the flexibility of the Retriever module allows practitioners to select the retrieval method and make further modification that best suits their task and data. The interface of \verb|Retriever| also allows users to pack those in-context examples and use them somewhere else. 

\paragraph{Inferencer}
\verb|Inferencer| invokes the pre-trained language model to generate predictions based on the concatenation of in-context examples and testing input. 
The \verb|Inferencer| supports various inference methods:
\begin{itemize}
    \item Direct: \citet{tom2020gpt3} use tokens in the vocabulary to represent candidate answers and select the final prediction using the one with the highest probability.
    \item Perplexity: ~\cite{tom2020gpt3} compute the sentence perplexity of the sequence concatenation of input and candidate answers and select the final prediction using the one with the lowest perplexity.
    \item Channel: \citet{min2022channel} proposed to utilize channel models~\cite{yu2016channel,  yee2019channel} to compute the conditional probability in a reversed direction, i.e., estimating the likelihood of input query given the label. 
\end{itemize}
The flexibility of \verb|Inferencer| also allows users to recursively invoke it to support multi-stage ICL methods, such as chain-of-thought~\citep{wei2022cot} and selection-inference~\citep{selectioninference}.
Additionally, \verb|Inferencer| can be augmented with a scorer to evaluate its prediction. 

\subsection{Efficiency}
To satisfy Principle P2, we equip OpenICL with various parallelism techniques to enable efficient inference for large-scale models. 
\paragraph{Data Parallel}
Data parallel~\citep{dataparallel} is a common technique used in parallel computing to improve the efficiency of large-scale computation tasks. OpenICL implements data parallelism to improve the performance of both the retrieval and inference steps. During  retrieval and inference, data is divided into smaller batches for processing. Additionally, for models that can fit into GPU's VRAM, OpenICL implements data parallelism by sharding the data across multiple GPUs and performing parallel inference on each GPU with a complete copy of the model. This significantly increases the inference speed when working with large datasets.

\paragraph{Model Parallel}
In the era of LLMs, models often have billions or hundreds of billions of parameters that exceed modern GPUs' capacity. To handle this problem, we resort to model parallel~\citep{modelparallel}: a parallel computing technique that divides a large deep learning model into smaller sub-models, each of which can be run on a separate GPU. 
OpenICL supports model parallelism that users can easily parallelize their models with minimal modification to the code. Currently, we support Megatron~\cite{megatron2019} and Zero~\cite{zero2019}.

\subsection{Generality}
To satisfy Principle P3, OpenICL is designed to maximize users' productivity by supporting a wide range of models, tasks, and methods:
\paragraph{[Model]} OpenICL supports both decoder-only LMs (e.g., GPT family~\cite{Radford2018ImprovingLU, radford2019language,gpt-neo,gpt-j,gptneox}, and encoder-decoder-based LMs (e.g., T5~\cite{2020t5}). We also provide two alternatives for accessing the model: users can directly load model checkpoints for evaluation or access a model via API (e.g., OpenAI's GPT-3 series models;~\citealt{tom2020gpt3,chen2021evaluating,ouyangtraining}).\footnote{\href{https://openai.com/api/}{https://openai.com/api/}}
\paragraph{[Tasks]} With the help of OpenICL, users can easily conduct experiments on both classification and generation tasks. OpenICL integrates HuggingFace's \verb|datasets|\footnote{\href{https://github.com/huggingface/datasets}{https://github.com/huggingface/datasets}} such that users can access and download thousands of NLP tasks with ease. 
\paragraph{[Methods]} As aforementioned, OpenICL provides broad support for ICL methods that cover both retrieval and inference. 
Furthermore, OpenICL offers the flexibility to return the results of the \texttt{Retriever} and \texttt{Inferencer} in a step-by-step manner, making it easy to integrate these intermediate results into other projects.

\section{Toolkit Walkthrough}
In this section, we demonstrate OpenICL by walking readers through several typical ICL use cases. 

\begin{figure*}[!thp]
\centering
\begin{minipage}{\linewidth}
\begin{lstlisting}[language=Python]
from openicl import DatasetReader, PromptTemplate
from openicl import TopkRetriever, PPLInferencer, AccEvaluator

# Load dataset
data = DatasetReader('gpt3mix/sst2', input_columns=['text'], output_column='label')  

# Define the prompt template for the task
tp_dict = { 0: '</E> Positive Movie Review: </Q>',
            1: '</E> Negative Movie Review: </Q>' }      
template = PromptTemplate(tp_dict, {'text':'</Q>'}, ice_token='</E>')

# Initiate the retriever and inferencer
retriever = TopkRetriever(data, ice_num=8)
inferencer = PPLInferencer(model_name='gpt2-xl')

# Run inference and calculate score
predictions = inferencer.inference(retriever, ice_template=template)
score = AccEvaluator().score(predictions=predictions, references=data.references)

\end{lstlisting}
\end{minipage}
\caption{Illustration of Example 1 which evaluates the ICL performance of GPT2-XL (1.5B) on SST-2 dataset with PPL inference strategy.}
\label{fig:code}
\end{figure*}

\begin{figure*}[!thp]
\centering
\begin{minipage}{\linewidth}
\begin{lstlisting}[language=Python]
from datasets import load_dataset 
from openicl import DatasetReader, PromptTemplate
from openicl import RandomRetriever, GenInferencer, BleuEvaluator

dataset = load_dataset("wmt16", 'de-en').map(lambda example: example['translation'])

data = DatasetReader(dataset, input_columns=['de'], output_column='en')

template = PromptTemplate('</E> German:</German> \n English: </English>',
                          {'de':'</German>', 'en':'</English>'}, ice_token='</E>')

retriever = TopkRetriever(data, ice_num=8)

# Inference by direct generation
inferencer = GenInferencer(model_name='facebook/xglm-7.5B')
predictions = inferencer.inference(retriever, ice_template=template)

# calculate Bleu
score = BleuEvaluator().score(predictions=predictions, references=data.references)

\end{lstlisting}
\end{minipage}
\caption{Illustration of Example 2 that evaluates the ICL performance of XGLM (7.5B) on WMT16 (de-en) dataset with direct inference strategy.}
\label{fig:code2}
\end{figure*}

\begin{figure*}[!thp]
\centering
\begin{minipage}{\linewidth}
\begin{lstlisting}[language=Python]
from openicl import DatasetReader, PromptTemplate, BM25Retriever, CoTInferencer

data = DatasetReader('gsm8k', name='main', 
                     input_columns=['question'], output_column='answer')

template = PromptTemplate('</E> Question: </Q> \n Answer: </A>',
                          {'question':'</Q>', 'answer':'</A>'},
                          ice_token='</E>')

retriever = TopkRetriever(data, ice_num=4)

# Inference by Chain-of-Thought
cot_list=["Let's think step by step.", 
          "\nTherefore, the answer (arabic numerals) is"]
          
inferencer = CoTInferencer(cot_list=cot_list, api_name='gpt3')
predictions = inferencer.inference(retriever, ice_template=template)
\end{lstlisting}
\end{minipage}
\caption{Illustration of Example 3, which evaluates the ICL performance of \textit{text-davinci-003} version of GPT-3 (175B) on GSM8K dataset with Chain-of-thought inference strategy.}
\label{fig:code3}
\end{figure*}

\paragraph{Example 1.} We first demonstrate how to use OpenICL to develop a typical ICL pipeline for language classification using a few lines of code and conduct evaluation on the popular sentiment classification dataset SST-2~\cite{sst2}. 
As shown in Figure~\ref{fig:code}, the pipeline begins with a \texttt{DatasetReader} which loads the dataset given its name on HuggingFace Dataset Hub\footnote{\href{https://huggingface.co/datasets}{https://huggingface.co/datasets}} or local file path. Users need to specify the names of columns where the input (``\textit{text}'') and output (``\textit{label}'') are stored. 
Secondly, a customized \texttt{PromptTemplate} is instantiated with a dictionary that defines the prompts for each class label. The placeholder </E> and </Q> will be replaced by in-context examples and testing input, separately. 
After that, we initiate the retriever based on TopK~\cite{liu2022makes} and set the number of in-context examples to 8 (``$ice\_num=8$''). We select perplexity-based method to initiate the inferencer and use GPT2-XL as the LLM. 
Having all these been set, we can run the inference by invoking the inferencer (line 17) and calculating the accuracy of the model's prediction(line 18). 

\paragraph{Example 2.} Figure~\ref{fig:code2} shows how to use OpenICL to work with generation problems. We consider the popular machine translation dataset WMT16~\citep{wmt16}. As in Example 1, we can easily load the dataset, define the prompt template, and initiate the retriever, by feeding new parameters to the function, respectively. The major API difference from Example 1 is that (i) we add some pre-processing for the translation task (line 5); (ii) \textit{PPLInferencer} is replaced by  inferencer tailored for generation (line 16); (iii) we use BLEU to evaluate model performance.

\paragraph{Example 3.} OpenICL also supports more advanced ICL methods, as shown in Figure~\ref{fig:code3}. Users can seamlessly switch to CoT by only modifying two lines of code: line 14 defines the template for CoT and line 15 initiates the inferencer with GPT3 using OpenAI's API.
Similar multi-step ICL methods such as Self-Consistency~\citep{selfconsistency} and Selection-Inference~\citep{selectioninference} can also be easily implemented by inheriting the superclass \texttt{Inferencer} designed in OpenICL.

\begin{figure*}[htbp]
    \centering
    \includegraphics[width=\textwidth]{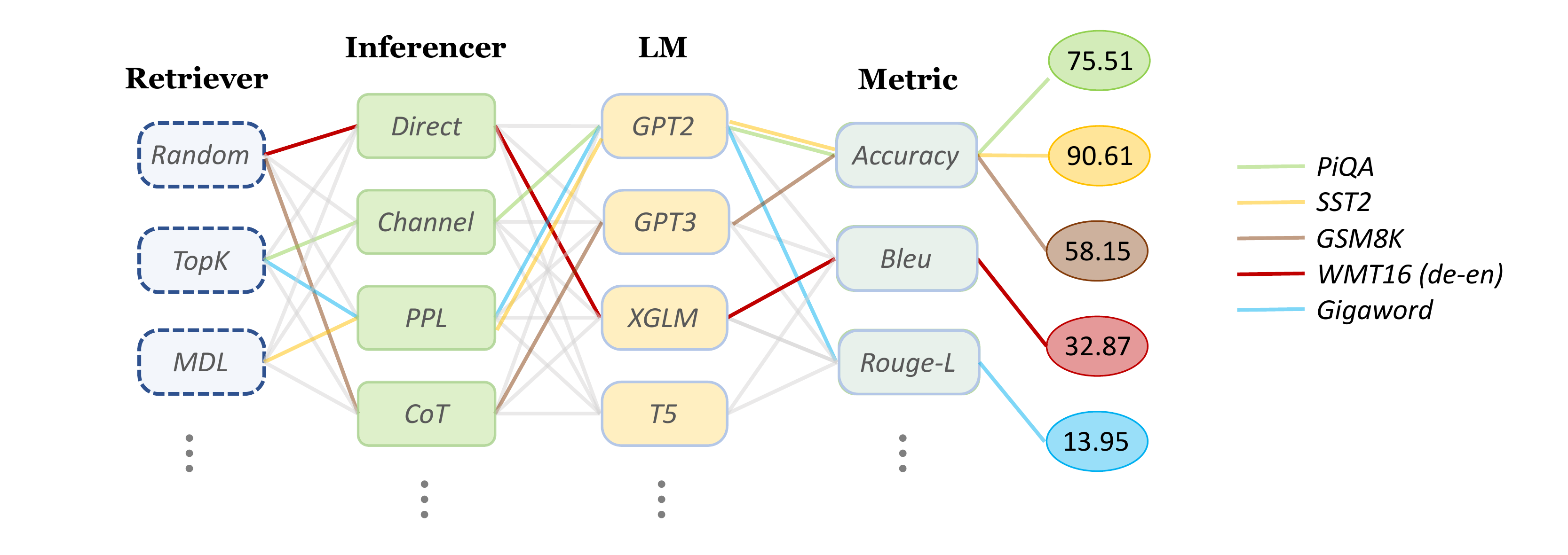}
    \caption{Evaluation results. We conduct experiments on five representative tasks with OpenICL and use different retrievers, inferencers, language models, and other components. In terms of model usage, we adopt GPT-Neo (2.7B) for SST2, PiQA, and Gigaword, XGLM (7.5B) for WMT16 (de-en), and \textit{text-davinci-003}  version of GPT-3 (175B) for GSM8K.}
    \label{fig:eval}
\end{figure*}
\section{Evaluation}
To demonstrate OpenICL's flexibility we conducted experiments on a diverse set of datasets, LLMs, and ICL methods. We consider  
PiQA~\citep{piqa} for commonsense reasoning, SST-2~\citep{sst2} for sentiment analysis, GSM8K~\citep{cobbe2021gsm8k} for arithmetic reasoning,  
WMT16 de-en~\citep{wmt16} for machine translation and Gigaword~\citep{gigaword} for summarization. 
We've also tested various LLMs, including GPT-Neo (2.7B)~\citep{gpt-neo,gao2020pile}, \textit{text-davinci-003} version of GPT-3 (175B), and XGLM (7.5B)~\citep{xglm}. We use OpenAI's official API\footnote{\href{https://openai.com/api/}{https://openai.com/api/}} to access GPT-3. The detailed setups and results are shown in Figure~\ref{fig:eval}.
As we can see, components of OpenICL can be easily chained to support different evaluation needs and replicate results of state-of-the-art methods.

\section{Conclusion}
We present OpenICL, an open-source toolkit for In-context learning. OpenICL provides a convenient and flexible interface for in-context learning practice and research. 
Our modular design allows it to support a wide range of LLMs, tasks, and ICL methods with ease. 
We implement both model parallelism and data parallelism to make inference of large models more efficient.
OpenICL is highly extensible, and we will continue to update it to keep pace with the latest research. Despite the promising results, ICL is still in its early stages, and many challenges remain. We believe OpenICL will be a valuable resource for researchers and practitioners to facilitate their research and development.

\bibliography{anthology,custom}
\bibliographystyle{acl_natbib}

\end{document}